\def\paperID{10036}
\def\confName{CVPR}
\def\confYear{2024}

\def\paperTitle{Feature Purified Transformer With Cross-level Feature Guiding Decoder For Multi-class OOD and Anomaly Deteciton}

\def\aauthor{Jerry Chun-Wei Lin\\
Silesian University of Technology\\
{\tt\small jerry.chun-wei.lin@polsl.pl}

\and
Pi-Wei Chen\\
National Cheng Kung University\\
{\tt\small nf6111015@gs.ncku.edu.tw}

\and
Chao-Chun Chen\\
National Cheng Kung University\\
{\tt\small chencc@imis.ncku.edu.tw}
}

\newif\ifreview 
\newif\ifarxiv \newcommand{\arxiv}{\arxivtrue}
\newif\ifcamera 
\newif\ifrebuttal 

\arxiv 

\pdfoutput=1
\documentclass[10pt,twocolumn,letterpaper]{article}
\ifreview \usepackage[review]{cvpr} \fi
\ifarxiv \usepackage[pagenumbers]{cvpr} \fi
\ifrebuttal \usepackage[rebuttal]{cvpr} \fi
\ifcamera \usepackage{cvpr} \fi

\usepackage{siunitx}
\usepackage{booktabs}
\usepackage{graphicx}	
\usepackage{amsmath}	
\usepackage{amssymb}	
\usepackage{booktabs}
\usepackage{times}
\usepackage{microtype}
\usepackage{epsfig}
\usepackage[table,xcdraw,dvipsnames]{xcolor}
\usepackage{caption}
\usepackage{float}
\usepackage{tabularray}
\usepackage{placeins}
\usepackage{color, colortbl}
\usepackage{stfloats}
\usepackage{enumitem}
\usepackage{tabularx}
\usepackage{xstring}
\usepackage{graphicx} 
\usepackage{multirow}
\usepackage{xspace}
\usepackage{url}
\usepackage{subcaption}
\usepackage{xcolor}
\usepackage[hang,flushmargin]{footmisc}

\ifcamera \usepackage[accsupp]{axessibility} \fi

% ...

\ifarxiv  \fi

\newcommand{\R}[1]{{%
    \textbf{%
        \ifstrequal{#1}{1}{\textcolor{red}{R#1}}{%
        \ifstrequal{#1}{2}{\textcolor{blue}{R#1}}{%
        \ifstrequal{#1}{3}{\textcolor{magenta}{R#1}}{%
        \ifstrequal{#1}{4}{\textcolor{teal}{R#1}}{%
                           \textcolor{cyan}{R#1}%
        }}}}%
    }%
}}

\usepackage{xr-hyper}

\makeatletter
\newcommand*{\addFileDependency}[1]{
  \typeout{(#1)}
  \@addtofilelist{#1}
  \IfFileExists{#1}{}{\typeout{No file #1.}}
}

\makeatother

\definecolor{cvprblue}{rgb}{0.21,0.49,0.74}
\usepackage[pagebackref,breaklinks,colorlinks,citecolor=cvprblue]{hyperref}
\usepackage[capitalize]{cleveref}
\crefname{section}{Sec.}{Secs.}
\crefname{table}{Table}{Tables}
\crefname{figure}{Fig.}{Figs.}

\begin{document}
\title{\paperTitle}
\author{\aauthor}
\maketitle

\begin{abstract}
Reconstruction networks are prevalently used in unsupervised anomaly and Out-of-Distribution (OOD) detection due to their independence from labeled anomaly data. However, in multi-class datasets, the effectiveness of anomaly detection is often compromised by the models' generalized reconstruction capabilities, which allow anomalies to blend within the expanded boundaries of normality resulting from the added categories, thereby reducing detection accuracy. We introduce the FUTUREG framework, which incorporates two innovative modules: the Feature Purification Module (FPM) and the CFG Decoder. The FPM constrains the normality boundary within the latent space to effectively filter out anomalous features, while the CFG Decoder uses layer-wise encoder representations to guide the reconstruction of filtered features, preserving fine-grained details. Together, these modules enhance the reconstruction error for anomalies, ensuring high-quality reconstructions for normal samples. Our results demonstrate that FUTUREG achieves state-of-the-art performance in multi-class OOD settings and remains competitive in industrial anomaly detection scenarios.
\end{abstract}

\section{Introduction}
Anomaly detection aims to identify outliers in a set of otherwise normal data and has wide-ranging applications in various domains such as cybersecurity \cite{evangelou2020anomaly}, healthcare \cite{vsabic2021healthcare}, and finance \cite{anandakrishnan2018anomaly}. One of the challenges in this area is the difficulty of obtaining labeled anomaly samples, which makes traditional supervised learning approaches \cite{gornitz2013toward} less feasible. As a result, unsupervised learning has become the prevalent paradigm for anomaly detection.
An unsupervised learning model trained with anomaly-free data learns the distribution of the normality feature. The model will identify any sample that deviates from the learned feature distribution as an anomaly, thereby achieving anomaly detection without explicit labels for anomalies.
\begin{figure}[!htbp]
\setlength{\abovecaptionskip}{0pt}
\setlength{\belowcaptionskip}{0pt} 
\centering 
   \includegraphics[width=\linewidth]{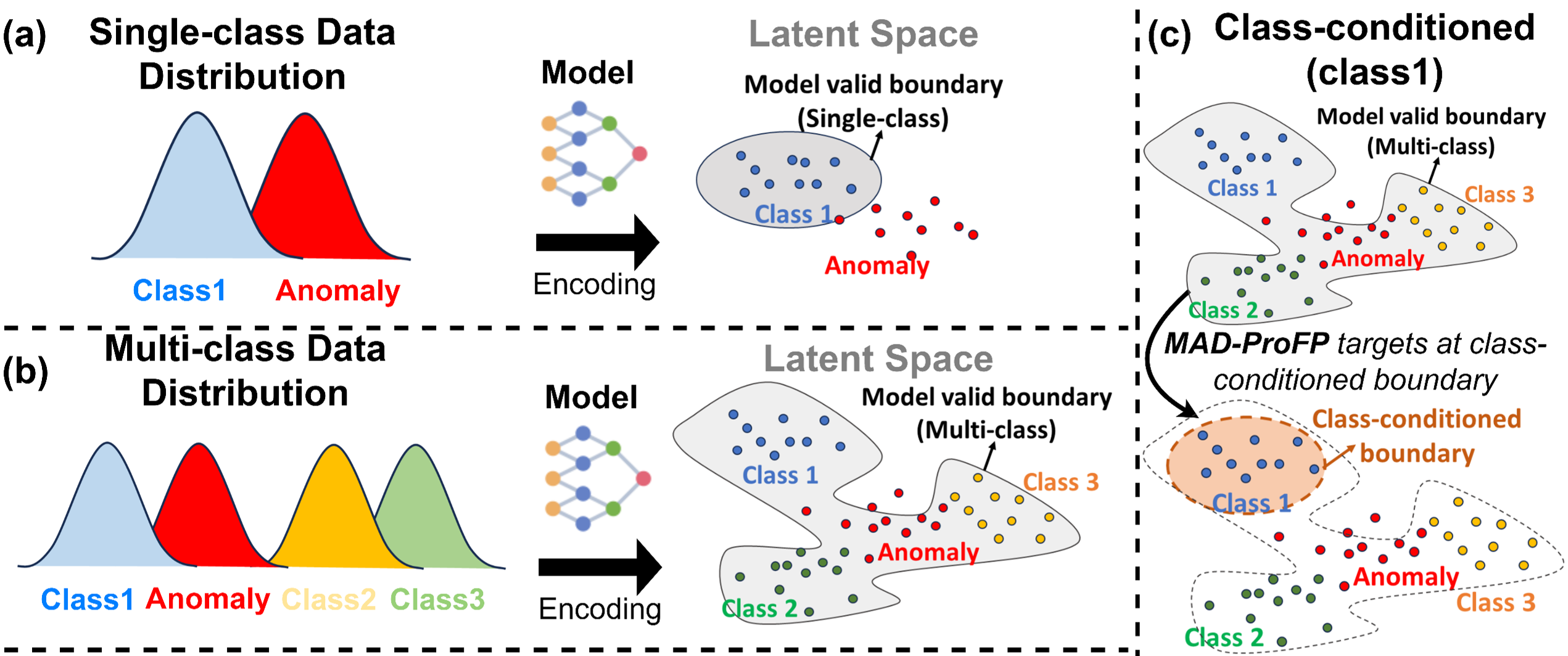}
   \caption{(a) illustrates the relationship between a single-class dataset and the latent space, specifically showing how the learned boundary excludes the anomalous embedding.; (b) illustrates the relationship between a multi-class dataset and the latent space, where the normality boundary tends to include anomalous embeddings; (c) illustrates our design philosophy that FUTUREG of constrain boundary for multi-class data into a class-conditioned boundary that applies exclusively to specific semantic classes in the latent space.}
\label{fig:1}

\end{figure}
Reconstruction networks (mostly autoencoder-based model)\cite{xie2016unsupervised,fan2018abnormal,schlegl2019f,lee2022anovit,an2015variational}
are widely used for unsupervised anomaly detection approach due to  their ability to capture the underlying features and distribution of normality from anomaly-free datasets.

The training process enables the model to disentangle the normality of the dataset into representative latent embeddings, thus capture the normality distribution. The training process involves encoding the sample into a latent variable and then decoding it back to its original input. The distribution of these latent embeddings acts as a boundary; embeddings that fall outside this boundary are considered as anomaly (as shown in Fig~\ref{fig:1}) and are
less likely to be accurately reconstructed, thereby inducing large reconstruction errors for anomalous samples. The reconstruction error is then used to distinguish anomalies by applying a defined threshold, which is typically determined based on the expected variation within normal data.

However, reconstruction networks face the risk of accurately reconstructing anomaly features, leading to a phenomenon known as 'identity shortcut'. This issue is particularly pronounced in a unified model within a multi-class setting \cite{lu2024hierarchical,you2022unified}, which complicates multi-class anomaly detection.

While previous works \cite{lu2024hierarchical,you2022unified} have leveraged query embedding in attention mechanisms to mitigate 'identity shortcuts' in industrial anomaly detection—where the focus is on identifying minor deviations within a homogeneous set of categories— these methods fall short in multi-class Out of Distribution (OOD) anomaly detection. In such scenarios, anomalies constitute a heterogeneous class, introducing significant difficulties in a multi-class setting due to anomalies being tied to class-level distinctions, which are further complicated by the increased diversity of data categories. Understanding the underlying causes of the 'identity shortcut' is imperative to excel across a broader spectrum of anomaly detection tasks.

In this work, we address the complex challenge of both industrial and Out of Distribution (OOD) anomaly detection in a many-vs-many setting. Our analysis reveals that 'identity shortcuts' largely stem from increased variability in the distribution boundaries of normality latent embeddings. This variability, when combined with the inherent diversity of multiple data categories, allows anomalous embeddings to more easily fall within these expanded boundaries, as illustrated in Figure \ref{fig:1}(b). Further statistical proofs provided in Section \ref{sec:Discussion of the behavior of reconstruction network in multi-class dataset} and Table.~\ref{table:statistacl1} and \ref{table:statistacl2} demonstrate that this leads to inaccurate reconstruction without significant errors. Consequently, it is essential to precisely refine the boundaries of normality latent embeddings to prevent the erroneous inclusion of anomalies, particularly those stemming from the heterogeneous classes characteristic of multi-class settings. Our primary motivation is to enhance the model's robustness against the increased variability and diversity, identified as the root causes of 'identity shortcuts' in industrial and OOD anomaly detection.

In this paper, we introduce a unified framework, the 'Feature pUrified Transformer with cross-level featURE Guiding Decoder (FUTUREG),' grounded in the Transformer architecture, which directly tackles the issue of 'identity shortcuts' stemming from increased variability in the distribution boundaries of latent representations. The core concept of FUTUREG is to dynamically constrain the original boundaries of latent embeddings to class-conditioned boundaries. This approach allows the model to effectively filter out potential anomalous embeddings while maintaining the original mapping functions, thereby preserving the model’s reconstruction capabilities without compromise. By refining the latent space in this manner, FUTUREG enhances the robustness of anomaly detection against the varied and diverse anomalies typically encountered in industrial and OOD settings.

To achieve our objective, we invent three main modules:  the ``Feature Purification Module (FPM)'' and ``, ``Normality Prototype Retrieval Module (NPRM)'', and 
Cross-level Feature Guiding (CFG) Decoder".
"The Feature Purification Module (FPM) is designed to constrain the expanded variability of latent variable boundaries through an attention-based similarity metric(detailed in Sec~\ref{sec:FPM}), which utilizes conditioned-prototypical\cite{snell2017prototypical} embeddings provided by the Normality Prototype Retrieval Module (NPRM)(detailed in Sec~\ref{sec:NPRM}).
To maintain the quality of reconstruction result, CFG decoder is design to use different level-encoder feature as query to guide the reconstruction process within a modified Transformer Decoder(detailed in Sec~\ref{sec:cfg}).

The FPM along  with NPRM and CFG Deocder is effective to restrict the over-expanded boundary of the normality latent embeddings without
compromising the model’s reconstruction capability.
The contribution of this paper can be summarized in three points:
\begin{itemize}
    \item We investigate the root cause of poor performance by reconstruction networks in multi-class datasets, providing a statistical explanation. Motivated by this limitation, we propose the 'FUTUREG' framework, designed for unified application across multi-class scenarios.
    \item We introduce the 'Feature Purification Module (FPM)', `NPRM' and the 'Cross-level Feature Guiding (CFG) Decoder' to address the challenges arising from increased variability in the boundary of latent embeddings which result in identity shortcut.
    \item We conduct comprehensive experiments to demonstrate the 'FUTUREG' framework's generalizability across various anomaly detection tasks. It achieves state-of-the-art (SOTA) results in multi-class OOD scenarios, leading by an average of 9\% and
    maintains comparable performance in industrial anomaly detection settings .
\end{itemize}
\section{Related Work}
\label{sec:related}

\subsection{Prototype learning}
Prototype learning is a specific approach to machine learning that focuses on learning representative instances, called prototypes, for each class within a feature space. One of the first applications of prototype learning in deep learning was learning with few shots, as \cite{snell2017prototypical} shows. The approach has also been used for segmentation tasks, as described by \cite{wang2019panet}.

Most existing methods train both feature representations and class prototypes in an end-to-end fashion. In general, prototypes start as learnable embeddings and provide a flexible framework that has been shown to be robust even when only a limited number of class examples are available, as well as to data variances or noise \cite{li2021adaptive}.
Recently, \cite{fontanel2021detecting} introduced prototype learning to the anomalous semantic segmentation domain, opening new avenues for its application in anomaly detection. A unique feature of prototype embedding is its centrality in latent space, effectively making it an anchor point around which all other class samples are arranged. This makes them invariant to small inter-class variances, a property we exploit in developing our framework for this work.

\subsection{Vision Transformer}
The transformer first proposed by \cite{vaswani2017attention} changed the way sequential data is processed. In the following year, \cite{dosovitskiy2020image} adopted Transformer in the computer vision field and showed better performance than CNN-based models in visual tasks.
The workflow involves splitting the image into a sequential embedding with an additional class token and passing it to the Transformer encoder for feature extraction. Finally, use an MLP head to perform classification on cls token.

Several studies \cite{mishra2021vt,lee2022anovit} adopted ViT in anomaly detection and show better performance than CNN-based models. However, its application is still limited to single-class datasets. You et al. \cite{you2022unified} discuss the suitability of Transformer in reconstruction-based anomaly detection and extend the application to a multi-class scenario by solving a phenomenon called ``identity truncation''. However, we argue that this is a suboptimal solution to the ``identity shortcut'' problem. Instead, in this paper, we explore the reasons for the ``identity shortcut'' and propose a framework to solve the fundamental problem and improve performance.

\section{Proposed Framework}
\label{sec:method}

\begin{figure*}
\setlength{\abovecaptionskip}{0pt}
\setlength{\belowcaptionskip}{0pt} 
\centering
\includegraphics[width=\linewidth]{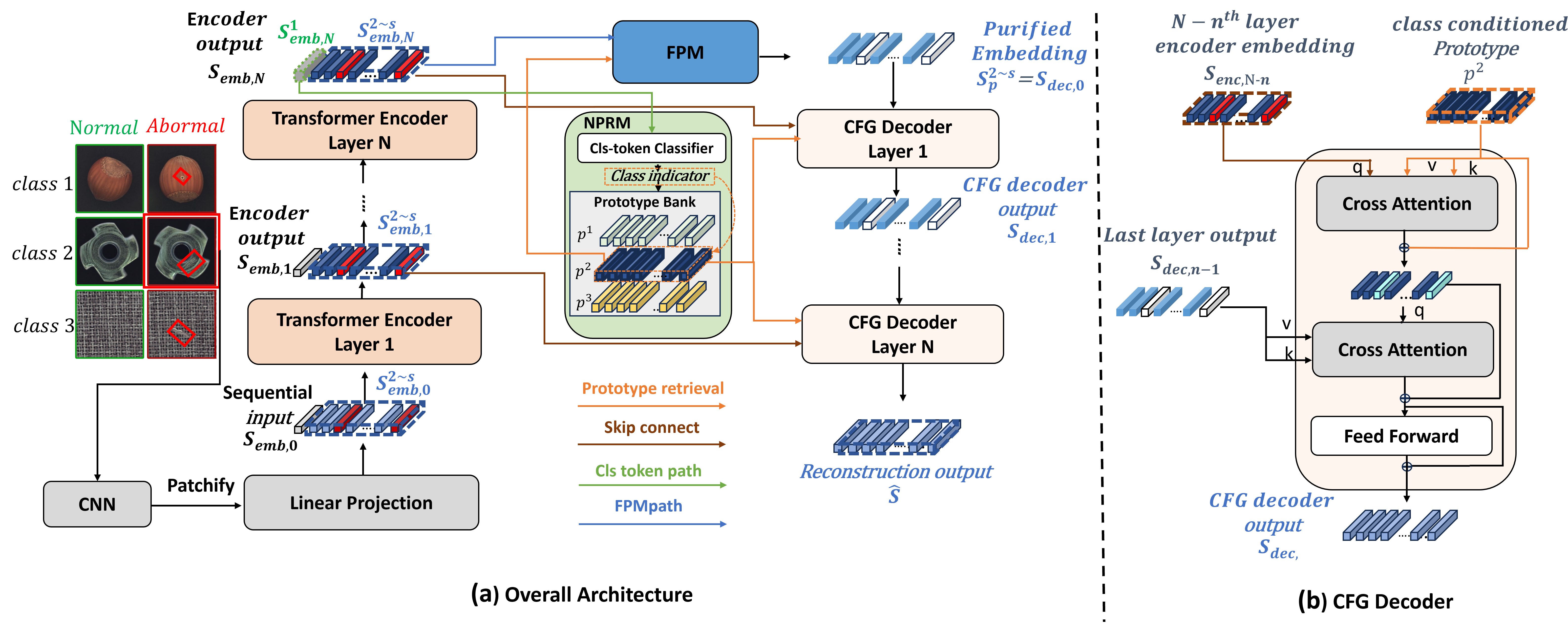}
   \caption{The overall architecture of FUTUREG}
\label{fig:Architecture}
\end{figure*}

\subsection{Overview}

The overall architecture of the FUTUREG framework is based on Transformer, and highlighted with the proposed Feature Purified Module (FPM) and Cross-level Feature Guiding (CFG) Decoder along with a Normality Prototype Retrieval Module (NPRM) as shown in Fig  ~\ref{fig:Architecture}.

The workflow of the proposed FUTUREG can be summarized into the following steps:
\begin{enumerate}
    \item[(i)] A CNN-based model processes the test sample $\mathbf{X}$, extracting feature maps that are then segmented into sequential patch tokens. These tokens, along with an additional classification token, form a sequence $\mathbf{S}$ represented in $\mathbb{R}^{(s+1) \times e}$, where $s$ is the sequence length.
    \item[(ii)] A vanilla Transformer encoder with $N$ layers transforms the sequential tokens into disentangled sequential embeddings $\mathbf{S_{emb,n}} \in \mathbb{R}^{(s+1) \times e}$ at each layer. This process is essential for refining the data's intrinsic features, enhancing the model’s accuracy in feature interpretation and anomaly detection.
    \item[(iii)] Within the NPRM, a Classifier uses the first classification token from the last encoder layer, $\mathbf{S_{emb,N}^1} \in \mathbb{R}^{1 \times e}$, to determine the sample's class $c$. It then selects the appropriate class prototype $p^{c}$ from a prototype bank and uses it along with patch sequential embeddings $\mathbf{S_{emb,N}^{2 \sim (s+1)}} \in \mathbb{R}^{s \times e}$ for refined feature extraction in the FPM module.
    \item[(iv)] The FPM module, anchored with the selected prototype $p^{c}$, refines the boundary to purify the last layer ($N$-th layer) sequential embeddings $\mathbf{S_{emb,N}^{2 \sim (s+1)}}$ from potential anomalies, producing a purified embedding $\mathbf{S_p^{2 \sim (s+1)}}$.
    \item[(v)] The CFG Decoder utilizes embeddings from each encoder layer $[\mathbf{S_{emb,1}^{2 \sim (s+1)}}, \mathbf{S_{emb,2}^{2 \sim (s+1)}}, \ldots, \mathbf{S_{emb,N}^{2 \sim (s+1)}}]$ and the corresponding prototype to guide the reconstruction of the final output $\mathbf{\hat{S}} \in \mathbb{R}^{s \times e}$, ensuring no loss of fine-grained features in the purified sequential embeddings.
\end{enumerate}

In the following subsection, we will present the mathematical analysis underlying the phenomenon of identity shortcuts, as detailed in Section~\ref{sec:Discussion of the behavior of reconstruction network in multi-class dataset}. This analysis highlights the motivations behind our methodological innovations. We will subsequently introduce the technical details of the Normality Prototype Retrieval Module (NPRM) in Section~\ref{sec:NPRM}, the Feature Purification Module (FPM) in Section~\ref{sec:FPM}, and the Cross-level Feature Guiding (CFG) Decoder in Section~\ref{sec:cfg}.
\subsection{The analysis of reason behind Identity shortcut }
\label{sec:Discussion of the behavior of reconstruction network in multi-class dataset}
In single-class anomaly detection, the reconstruction algorithm projects normal samples into a latent space defined by a conceptual boundary. This boundary enables perfect reconstruction of normal samples, while anomalies, falling outside this boundary, induce significant errors. However, in multi-class scenarios, increased variability among classes expands this boundary to incorporate data points that should be anomalies. This expansion, known as `identity shortcut,' dilutes the model’s ability to distinguish between normal and anomalous data, as anomalies are reconstructed with fewer errors and treated as normal occurrences. We present statistical evidence to substantiate the issue of an overly expansive boundary in multi-class anomaly detection settings.

Consider modeling the latent space of a single-class dataset \(X_1\) with a Gaussian distribution characterized by a mean \(\mu_1\) and a variance \(\sigma_1^2\):
\begin{equation}
p(X_1) \sim \mathcal{N}(\mu_1, \sigma_1^2)
\end{equation}
In this model, the Gaussian parameters \(\mu_1\) and \(\sigma_1^2\) define a statistical boundary for normalcy within the latent space. This boundary is conceptual, representing areas where data samples are most likely to be found, as dictated by the high probability density under the Gaussian curve.

The probability density function (PDF) for this distribution is given by:
\begin{equation}
    f_1(X) = \frac{1}{\sqrt{2\pi\sigma_1^2}} \exp\left(-\frac{(X - \mu_1)^2}{2\sigma_1^2}\right)
\end{equation}
This function illustrates how the probability decreases as \(X\) deviates from \(\mu_1\), effectively highlighting how anomalies, which fall in the lower probability tails, are identified.

Introducing a second class of data, \(X_2\), to the dataset necessitates modeling the combined data with a Gaussian distribution characterized by a mean \(\mu_2\) and a larger variance \(\sigma_2^2\) to accommodate the increased diversity:
\begin{equation}
p(X_2) \sim \mathcal{N}(\mu_2, \sigma_2^2), \quad \sigma_2^2 > \sigma_1^2.
\end{equation}

The probability density function (PDF) for this two-class distribution is:
\begin{equation}
f_2(X) = \frac{1}{\sqrt{2\pi\sigma_2^2}} \exp\left(-\frac{(X - \mu_2)^2}{2\sigma_2^2}\right)
\end{equation}

A higher PDF value indicates a greater likelihood of a sample \(X\) falling within the normal distribution boundary defined by \(\mu_2\) and \(\sigma_2^2\). Consider an outlier sample \(x_{\text{test}}\) that exhibits approximately the same displacement from the means (\(\mu_1\) and \(\mu_2\)) of the single-class and multi-class distributions:
\begin{equation}
|X_{\text{test}} - \mu_1| \approx |X_{\text{test}} - \mu_2|
\end{equation}
Given the larger variance \(\sigma_2^2\) compared to \(\sigma_1^2\), the outlier sample \(x_{\text{test}}\) is expected to yield a higher PDF value in the two-class distribution, indicating reduced sensitivity to anomalies. This is because the larger variance \(\sigma_2^2\) results in a slower decay rate in the exponential term, leading to a higher probability value, especially due to the exponential function’s effect:
\begin{equation}
\exp\left(-\frac{(X_{\text{test}} - \mu_1)^2}{2\sigma_1^2}\right) < \exp\left(-\frac{(X_{\text{test}} - \mu_2)^2}{2\sigma_2^2}\right)
\end{equation}

This mathematical behavior implies that in the presence of a larger variance, outliers are less sharply penalized, reflecting a broader tolerance within the distribution and suggesting that the outlier \(x_{\text{test}}\) is more likely to be considered within the region of normal sample distribution boundary. Further experimental proof is shown in Table.~\ref{table:statistacl1} and \ref{table:statistacl2}.

\subsection{Normality Prototype Retrieval Module}
\label{sec:NPRM}
Based on the the observation elaborated in Sec~\ref{sec:Discussion of the behavior of reconstruction network in multi-class dataset}, we are inspired to leverage the prototype learning to estabilshed a prototype bank $\{p^{1}_{\theta}, p^{2}_{\theta}, \ldots, p^{T}_{\theta} \}, p^{i}_{\theta} \in \mathbb{R}^{s \times e}$ in which each prototype embedding $p^{i}$ is encoded  with normality semantic of each class item during training which only normal sample are engaged, which enable us to anchor the latent space to form a dynamic boundary that restrict normality- irrelevant embedding in the latent space.

Within the NPRM, A Cls-token Classifier which composed of a simple FC network $f_c()$ takes the cls token embedding $\mathbf{S_emb^1}$ and output $\mathbf{y}$ indicating  the probability of the input $\mathbf{X}$ belonging to a specific class:
\begin{equation}
    \mathbf{y} = f_c(S^1_{\text{emb}})
\end{equation}
Where each element $y_i$ of vector $\mathbf{y}$ corresponds to the probability that the input $\mathbf{X}$ belongs to class $i$.
The prototype embedding $p^{c}$ is selected based on the most probable class $c$ where c is :
\begin{equation}
    c = \arg\max_i y_i
\end{equation}
The selected prototype can be further utilized for fusion with the encoder embedding, and serve as complementary feature to guide the reconstrucition.

\textbf{Prototype Bank:}
A set of learnable embedding $\{p^{1}_{\theta}, p^{2}_{\theta}, \ldots, p^{T}_{\theta} \}, p^{i}_{\theta} \in \mathbb{R}^{s \times e}$ is trained with anomaly-free data in the training phase optimized to capture the semantic essence of each class in an anomaly-free context. \( T \) is the number of classes, \( s \) is the sequence length of the embedding, and \( e \) is the dimension of the embedding.
Once trained, each prototype $p^{i}_{\theta}$ is fixed, thereby preserving the integrity of the semantic information for each class.

\subsection{Feature Purification Module}  
\label{sec:FPM}
\begin{figure*}
\setlength{\abovecaptionskip}{0pt}
\setlength{\belowcaptionskip}{0pt} 
\centering 
\includegraphics[width=\linewidth]{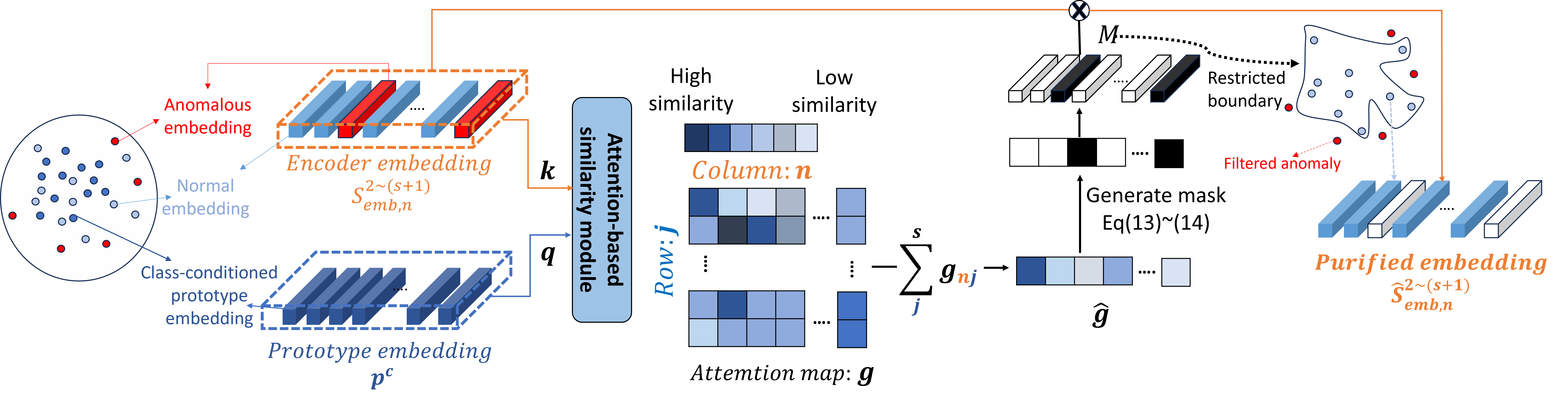}
   \caption{(a) illustrates the workflow of the proposed FPM; (b) illustrates the concept of how top-$k$ selection allows us to from a selective boundary that eliminates potential anomalous embedding.}
\label{FPM}
\end{figure*}

FPM plays a crucial role in avoiding over-expanded boundaries by establishing a restrictive boundary in the latent space. This effectively excludes anomalous embeddings during testing by masking those embeddings that do not align with the prototype embedding \( p^c \). This approach partitions the general latent space into a class-conditioned subspace, addressing the challenge discussed in Section~\ref{sec:Discussion of the behavior of reconstruction network in multi-class dataset}.

To establish this restrictive boundary, the selection of the \( k \) most relevant embeddings from the encoder representation \( \mathbf{S_{emb,n}^{2\sim(s+1)}} \), corresponding to the chosen prototype \( p^c \), forms a selective boundary in the latent space, as depicted in Figure~\ref{FPM}(a). This model focuses on embeddings that align with the corresponding semantic region, offering robustness against semantic-level variations. An attention-based similarity measure is utilized to calculate the similarity matrix \( \mathbf{g} \) between the last encoder representation \( \mathbf{S_{emb,n}^{2\sim(s+1)}} \) and the class-conditioned prototype \( p^c \) by their corresponding derived embeddings \( k_h \) and \( q_p \):

\begin{equation}
\mathbf{g} = \text{Softmax}\left(\frac{q_p \cdot k_h^\top}{\sqrt{d}}\right), \quad \mathbf{g} \in \mathbb{R}^{s \times s}
\end{equation}

where \( q_p \) is the query embedding derived from \( \mathbf{S_{emb,n}^{2\sim(s+1)}} \), with dimensions \( \mathbb{R}^{s \times d} \) (representing \( s \) sequences, each of dimension \( d \)); \( k_h \) is the matrix of key embeddings also derived from \( \mathbf{S_{emb,n}^{2\sim(s+1)}} \), with identical dimensions \( \mathbb{R}^{s \times d} \). The similarity matrix \( \mathbf{g} \) provides information on how much each encoder embedding from \( \mathbf{S_{emb,n}^{2\sim(s+1)}} \) matches the "normality" represented by the class-conditioned prototype \( p^c \). To gauge the overall similarity of each element from \( \mathbf{S_{emb,n}^{2\sim(s+1)}} \) to all elements in \( p^c \) within the latent space, the rows of \( \mathbf{g} \) are summed to yield an average row similarity \( \hat{g} \in  \mathbb{R}^{s} \):

\begin{equation}
    \hat{g}_n = \sum_{j=1}^{s} g_{nj}
\end{equation}

Instead of directly selecting the top-\( k \) most similar embeddings from \( \mathbf{S_{emb,n}^{2\sim(s+1)}} \), a masking approach is applied where embeddings not belonging to the top-\( k \), based on the similarity scores in \( \hat{g} \), are excluded. The top-\( k \) indices \( I_k \) are determined by sorting \( \hat{g} \) in descending order and selecting the first \( k \) indices:

\begin{equation}
I_k = \text{argsort}(-\hat{g})[:k]
\end{equation}

To mask the \( k \) remaining embeddings, an indicator vector \( \mathbf{1}_{I_k} \) specifies the value at the \( i \)-th position of the vector:

\begin{equation}
\mathbf{1}_{I_k}(i) = 
    \begin{cases} 
        1 & \text{if } i \in I_k, \\
        0 & \text{otherwise}
    \end{cases}
\end{equation}

This indicator vector is then transformed into a diagonal matrix \( \mathbf{M} \), enabling the matrix to function as a mask when applied to \( \mathbf{S_{emb,n}^{2\sim(s+1)}} \):

\begin{equation}
\mathbf{M} = \text{diag}(\mathbf{1}_{I_k})
\end{equation}

\begin{equation}
\mathbf{\hat{S_{emb}^{2\sim(s+1)}}} = \mathbf{M} \cdot \mathbf{S_{emb,n}^{2\sim(s+1)}}
\end{equation}

The above process, illustrated in Figure~\ref{FPM}(b), allows us to retain the top \( k \) similar embeddings that align with the normality feature prototype \( p^c \), thereby forming a more compact boundary of the class feature, avoiding an over-expanded feature distribution.

\subsection{Cross-level Feature Guiding (CFG) Decoder}
\label{sec:cfg}
The CFG Decoder is designed to decode the purified embedding \( \mathbf{\hat{S}_{emb}^{2\sim(s+1)}} \) back to the original input without reconstructing the anomalous features. The FPM filters out the \( s-k \) embeddings, potentially excluding detailed features. To counter this, we utilize a dual cross-attention mechanism in the CFG Decoder that integrates various levels of encoder outputs with the prototype \( p^c \), thus enriching the purified embeddings by integrating detailed information from multiple abstraction levels.

For $i$-th layer of CFG decoder performs two sequential cross-attention operations:
\begin{enumerate}
    \item The first operation fuses the corresponding encoder layer (e.g, the corresponding encoder layer for first layer of CFG decoder is the last layer encoder, which can be calculated by $N-i+1$, where $N$ is the total number of layer and $i$ is the current layer) output \( \mathbf{S}_{emb, n-i+1}^{2\sim(s+1)} \) with the prototype \( p^c \), using \( p^c \) as both the key and value, to produce an intermediate fused feature \( \mathbf{F}_i^{inter} \).
    \item The second operation refines this feature by attending to the output of the previous CFG layer (\( \mathbf{S}_{dec, i-1} \)), using \( \mathbf{F}_i^{inter} \) as the query.
\end{enumerate}

The attention processes for both layers are defined as follows:

\begin{equation}
\begin{split}
\mathbf{V}_i &= \mathbf{W}_v p^c, \\
\mathbf{K}_i &= \mathbf{W}_k p^c, \\
\mathbf{Q}_i &= \mathbf{W}_{q} \mathbf{S}_{emb, N-i+1}^{2\sim(s+1)}
\end{split}
\end{equation}

\begin{equation}
\mathbf{F}_i^{inter} = \text{Softmax}\left(\frac{\mathbf{Q}_i \mathbf{K}_i^\top}{\sqrt{d_k}}\right) \mathbf{V}_i
\end{equation}

For the second attention mechanism within each CFG layer:

\begin{equation}
\begin{split}
\mathbf{Q}_{i,2} &= \mathbf{W}_{q,2} \mathbf{F}_i^{inter}, \\
\mathbf{K}_{i,2} &= \mathbf{W}_{k,2} \mathbf{S}_{dec, i-1}, \\
\mathbf{V}_{i,2} &= \mathbf{W}_{v,2} \mathbf{S}_{dec, i-1}
\end{split}
\end{equation}

\begin{equation}
\mathbf{S}_{dec, i} = \text{Softmax}\left(\frac{\mathbf{Q}_{i,2} \mathbf{K}_{i,2}^\top}{\sqrt{d_k}}\right) \mathbf{V}_{i,2}
\end{equation}

Here, \( \mathbf{W}_v \), \( \mathbf{W}_k \), \( \mathbf{W}_{q} \), \( \mathbf{W}_{v,2} \), \( \mathbf{W}_{k,2} \), and \( \mathbf{W}_{q,2} \) are trainable parameters transforming the prototype \( p^c \), the intermediate features, and the previous decoder outputs into their respective representations in the value, key, and query spaces for both attention stages. The dimension \( d_k \), typically the square root of the key's dimension, is used as a scaling factor to stabilize the gradients during training.

The output of the last CFG decoder layer \( \mathbf{S}_{dec, N} \) represents the reconstructed output \( \mathbf{\hat{S}} \in \mathbb{R}^{s \times e} \), ensuring that the final reconstructed input is devoid of anomalous characteristics yet retains essential details. This layered attention mechanism allows for refined control over the feature integration process, ensuring that each layer's output is conditioned appropriately on the inputs and the ongoing reconstruction process.

\subsection{Objective Function}
Our objective function is designed to optimize two main aspects: accurate classification of the input object and high fidelity in reconstructing the input embeddings. These components are balanced in a weighted sum to form the overall loss, which guides the training of the network.

\textbf{Classification Loss}
The classification loss is used to accurately predict the category \(\hat{y}\) of the input object by minimizing the cross-entropy between the ground truth labels \(\mathbf{\hat{y}}\) and the predicted probabilities \(\mathbf{y}\). The cross-entropy loss for a multi-class classification problem is defined as:
\begin{equation}
    L_{cls} = -\sum_{i} \hat{\mathbf{y}}_i \log(\mathbf{y}_i),
\end{equation}
where \(\mathbf{\hat{y}}\) is the one-hot encoded ground truth vector, and \(\mathbf{y}_i\) is the predicted probability of the \(i\)-th class. This loss function is critical for training the model to produce outputs that align closely with the actual category labels, thereby enhancing classification accuracy.

\textbf{Reconstruction Loss}
The primary objective of the Reconstruction network is to minimize the error between the original input embeddings and the reconstructed outputs. Unlike typical approaches that use raw pixel data, our approach focuses on reconstructing the patch sequential embeddings \(\mathbf{S_{emb}^{2\sim(s+1)}}\) that are processed prior to the Transformer. These embeddings are represented as \(\mathbf{S_{emb}^{2\sim(s+1)}} \in \mathbb{R}^{s \times e}\), and the reconstructed output is denoted by \(\mathbf{\hat{S}} \in \mathbb{R}^{s \times e}\). The reconstruction loss is computed as follows:
\begin{equation}
    L_{rec} = \frac{1}{s} \sum_{i=1}^{s} || \mathbf{S}_{emb, i}^{2\sim(s+1)} - \mathbf{\hat{S}_i} ||_2^2,
\end{equation}
where \(\mathbf{S}_{emb, i}^{2\sim(s+1)}\) is the \(i\)-th embedding vector in the sequence of original embeddings, and \(\mathbf{\hat{S}_i}\) is the corresponding reconstructed embedding vector. This loss function encourages the network to preserve essential details in the embeddings, thereby aiding in better recovery of the input data while filtering out noise and anomalies.

\textbf{Overall Objective Function}
The overall objective function \(L\) integrates both classification and reconstruction objectives, allowing the model to balance between accurate classification and high-fidelity reconstruction:
\begin{equation}
    L = (1-\alpha)L_{rec} + \alpha L_{cls},
\end{equation}
where \(\alpha\) is a hyperparameter that balances the relative importance of the reconstruction loss \(L_{rec}\) and the classification loss \(L_{cls}\). Adjusting \(\alpha\) enables the model to prioritize either classification accuracy or reconstruction fidelity depending on the specific requirements of the task.

This balanced approach ensures that the model does not overly focus on one aspect at the expense of another, potentially leading to more robust and generalizable performance across different types of input data. 

\section{Experiments}
\label{sec:experiment}
\begin{figure*}
\setlength{\abovecaptionskip}{0pt}
\setlength{\belowcaptionskip}{0pt} 
\centering
\includegraphics[width=\linewidth, height=0.7\linewidth, keepaspectratio]{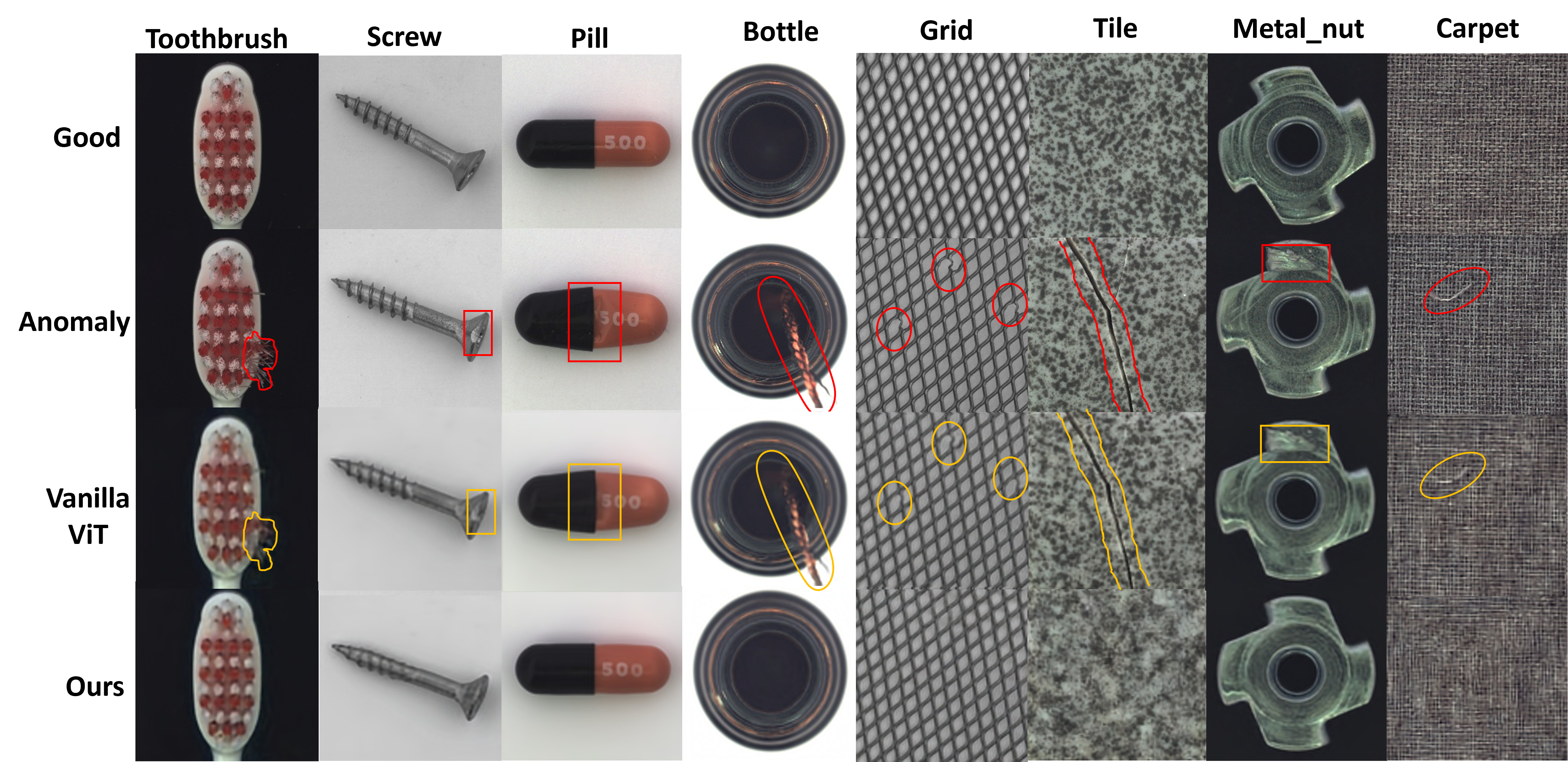}
\caption{The visualization of the comparison result between vanilla ViT and the proposed framework. Our proposed framework converts the anomalous pixels into normality-like pixels, causing a larger reconstruction error for the anomalous sample. The first and second rows are the image of an anomaly-free item and an anomalous item respectively. The third and fourth rows are the visualization of the reconstruction embedding of UniAD and our proposed MAD-ProFP.}
\label{MVtec_recon}
\end{figure*}

In this work, we test our proposed MAD-ProFP on CIFAR10 \cite{krizhevsky2009learning} and Mnist \cite{deng2012mnist} for validating it's efficacy on
multi-class Out-of-Distribution (OOD) detection.
CIFAR 10 and Mnist are two benchmark datasets with 10 classes for the classification task. We randomly select 5 categories as normal and set the remaining categories as anomalous (e.g., \{01234\} means that we select the first 5 categories as normal and the remaining are defined as anomalous, which do not appear in our training time). Unlike MVTec, we try to identify an unseen class (semantic shift) in the testing phase.

Furthermore, We also include MVTec-AD \cite{bergmann2019mvtec} dataset to showcase its robustness in anomaly detection within industrial scenario.
MVTec is a benchmark dataset for industrial anomaly detection task
and consists of 15 different categories. Each category has 5-6 subcategories with different defect types.
Different from OOD detection, in this dataset, we aim to identify anomalaus feature within homogeneous item.

We treat the above anomaly detection task in a many-vs-many environment where the model is trained with joint normal dataset.
We use the Area Under Region of Curve (AUROC) to evaluate the model to eliminate the threshold factor for the reconstruction error.

\subsection{Implementation details }
We resize all input images to a size of 224x224x3, followed by the use of a pretrained EfficientNet\cite{tan2019efficientnet} to extract a 14x14x272 feature map. The channel size is then reduced to 256 through a linear transformation. The patch size is set at 16 for segmenting the feature map into sequences. We employ the AdamW optimizer [citation needed] with a weight decay of 0.0001. The training process spans 1,000 epochs with an initial learning rate of 0.0001, which is decreased by a factor of 0.1 at epoch 800.
\subsection{OOD tasks comparison}
As shown in Table \ref{tab:Mnist_comparison}, our proposed method shows significant performance improvements over previous SOTA on OOD detection tasks; specifically, we outperform HVQ-Trans by 8.4\% and 5.6\% on CIFAR10 and MNIST, respectively. We argue that although previous unified methods (\cite{you2022unified,lu2024hierarchical}) address the identity shortcut issue in industrial scenarios, they still fail to address the core problem of over-spanned reconstruction boundaries in the latent space, which occurs specifically in many-vs.-many OOD settings. The proposed FUTUREG is designed to prevent over-spanned reconstruction boundaries specifically, thereby showing significant improvement over previous SOTA methods.
\begin{table*}[!ht]
\caption{AUROC performance comparison on multi-class OOD with MNIST and CIFAR10 datasets.} 
\label{tab:Mnist_comparison}
 \resizebox{\textwidth}{!}{%
\begin{tabular}{lcccccccc}
\toprule
\multicolumn{8}{c}{\textbf{AUROC performance on MNIST and CIFAR10}} \\
\toprule
Normal Indices & US\cite{bergmann2020uninformed} & FCDD\cite{yi2020patch} & FCDD+OE\cite{yi2020patch} & PANDA\cite{reiss2021panda} & MKD\cite{salehi2021multiresolution} & UniAD\cite{you2022unified} & HVQ-Trans\cite{lu2024hierarchical}& ours \\
\midrule
\{01234\} & 50.8/51.3 & 53.2/55.0 & 68.5/71.8 & 68.3/66.6 & 67.5/64.2 & 81.4/84.4 & 81.5/82.6 &98.1/93.3  \\
\{56789\} & 52.3/51.3 & 51.5/50.3 & 69.5/73.7 & 65.3/73.2 & 67.3/69.3 & 81.2/80.9 & 80.4/84.3&93.2/ 89.9  \\
\{02468\} & 51.5/63.9 & 53.4/59.2 & 64.5/85.3 & 67.5/77.1 & 71.2/76.4 & 80.3/89.3 & 82.4/92.4& 85.3/97.2 \\
\{13579\} & 55.2/56.8 & 54.0/58.5 & 74.8/85.0 & 73.2/72.9 & 68.9/78.7 & 80.3/90.6 & 83.2/91.9& 84.5/93.4 \\
\midrule
Mean & 52.4/55.9 & 53.0/55.8 & 69.3/78.9 & 68.5/72.4 & 68.7/72.1 & 80.8/87.2 & 81.8/87.8 &90.2/93.4 \\
\bottomrule
\end{tabular}}
\end{table*}
\subsection{Image-level and pixel-level industrial anomaly comparison}


As shown in Table \ref{tab:MVTec_updated}, our approach, FUTUREG, exhibits robust anomaly detection capabilities, achieving impressive AUROC scores of 96.3/96.6\% (image level/pixel level). These results are closely competitive with the top scores of 98.0/97.3\% achieved by HVQ-Trans, illustrating a marginal performance difference of only 1-2\%. This demonstrates the high efficacy of our model in delivering near-SOTA results in a challenging benchmark.

Furthermore, FUTUREG’s superior performance in industrial applications and its leadership in multi-class OOD scenarios reveal its broader, more versatile capabilities. These achievements highlight FUTUREG’s unique contributions and innovative approach compared to previous works, particularly in its ability to handle complex open-set recognition\cite{geng2020recent}.

 
\begin{table*}[!ht]

\centering
\caption{Anomaly detection/localization results with AUROC metric on MVTec-AD. All methods are evaluated under the many v.s many settings. The learned model is applied to detect anomalies for all categories without fine-tuning. The best results are bold with black.}
\label{tab:MVTec_updated}
\resizebox{\textwidth}{!}{
\scriptsize 
\begin{tabular}{l|S[table-format=2.1]|S[table-format=2.1]|S[table-format=2.1]|S[table-format=2.1]|S[table-format=2.1]|S[table-format=2.1]|S[table-format=2.1]|S[table-format=2.1]|S[table-format=2.1]|S[table-format=2.1]|S[table-format=4.2]}
\toprule
Category & \multicolumn{1}{c|}{PSVD\cite{yi2020patch}} & \multicolumn{1}{c|}{PaDiM\cite{defard2021padim}} & \multicolumn{1}{c|}{MKD\cite{salehi2021multiresolution}} & \multicolumn{1}{c|}{DRAEM\cite{zavrtanik2021draem}} & \multicolumn{1}{c|}{SimpleNet\cite{liu2023simplenet}} & \multicolumn{1}{c|}{PatchCore\cite{roth2022towards}} & \multicolumn{1}{c|}{RD4AD\cite{deng2022anomaly}} & \multicolumn{1}{c|}{UTRAD\cite{chen2022utrad}} & \multicolumn{1}{c|}{UniAD\cite{you2022unified}} & \multicolumn{1}{c|}{HVQ-Trans\cite{lu2024hierarchical}} & \multicolumn{1}{c}{Ours} \\
\midrule
Bottle    & {85.5/86.7} & {97.9/96.1} & {98.7/91.8} & {97.5/87.6} & {97.7/91.2} & {100/97.4} & {98.7/97.7} & {100/96.4}&{99.7/98.1} & {100/100} & {100/97.7} \\
Cable    & {64.4/62.2} & {70.9/81.0} & {78.2/89.3} & {57.8/71.3} & {87.6/88.1} & {95.3/93.6} & {85.0/83.1} & {97.8/97.1} &{95.2 / 97.3}& {99.0 /98.1} & {88.1/95.8} \\
Capsule    & { 61.3/83.1} & {73.4/96.9} & {68.3/88.3} & {65.3/50.5} & {78.3/89.7} & {96.8/98.0} & { 95.5/98.5} & {82.0/97.2} & { 86.9 / 98.5} & {95.8/98.8}& {89.6/98.5} \\
Hazelnut   & {83.3/97.4} & {85.5/96.3} & {97.1/91.2} & {93.7/96.9} & {99.2/95.7} & {99.3/97.6} & {87.1/98.7} & {99.8/98.2} & {99.8/98.1} & {100/98.8} & {99.9/97.8}\\
Metal Nut  & {80.9/96.0} & {88.0/84.8} & {64.9/64.2} & {72.8/62.2} & {85.1/90.9} & {99.1/96.3} & {99.4/94.1} & {94.7/96.4} & {99.2/94.8} & {99.9/96.3} &{99.2/94.9} \\
Pill       & {89.4/96.3} & {68.8/87.7} & {79.7/69.7} & {82.2/94.4} & {78.3/89.7} & {86.4/96.5} & {52.6/96.5} & {89.7/95.7} & {93.7/95.0} & {95.8/97.1}& {92.7/95.7} \\
Screw      & {80.9/74.3} & {56.9/94.1} & {75.6/92.1} & {82.0/94.5} & {45.5/93.7} & {94.2/98.9} & {97.3/99.4} & {75.1/95.2} & {87.5/98.3} & {95.6/98.9 }& {88.9/98.5} \\
Toothbrush & {99.4/98.0} & {95.3/95.6} & {75.3/88.9} & {90.6/97.7} & {94.7/97.5} & {100/99.0} & { 99.4/99.0} & {89.7/94.0} & {94.2/98.4} & {93.6/98.6}& {92.7/98.1} \\
Transistors & {77.5/95.1} & {86.6/92.3} & {73.4/71.7} & {74.8/64.5} & {82.0/86.0} & {98.9/92.3} & {92.4/86.4} & {92.0/91.5} & {98.9/97.9} & {99.7/97.9} & {99.2/97.1}\\
Zipper     & {77.8/95.1} & {79.7/94.8} & {87.4/86.1} & {98.8/98.3} & {99.1/97.0} & {97.1/95.7} & {99.6/98.1} & {95.5/97.3} & {95.8/96.8 } & {97.9/97.5}& {97.6/97.5} \\
Carpet    & {63.3/78.6} & {93.8/97.6} & {69.8/95.5} & {98.0/98.6} & {95.9/92.4} & {97.0/98.1} & {97.1/98.8} & {80.3 / 94.4} & {99.8/98.5} & {99.9/98.7}&{99.7/98.3} \\
Grid       & {66.0/70.8} & {73.9/71.0} & {83.8/82.3} & {99.3/98.7} & {49.8/46.7} & {91.4/99.2} & {99.7/99.2} & {93.9 / 95.2} & {98.2/96.5} & {97.0/97.0}& {98.3/97.0} \\
Leather    & {60.6/93.5} & {99.9/84.8} & {93.1/91.2} & {99.8/98.0} & {93.9/96.9} & {100/99.4} & {100/99.4} & {99.8 / 98.4} & {100/98.4} & {100/98.8} & {100/98.5} \\
Tile       & {88.3/92.1} & {93.3/80.5} & {89.5/85.3} & {99.8/98.0} & {93.7/93.1} & {96.0/90.3} & {97.5/95.6} & {98.8 / 94.2} & {99.3/91.8} & {99.2/99.2}& {99.4/91.2} \\
Wood       &{72.1/80.7} & {98.4/89.1} & {93.4/80.5} & {99.8/96.0} & {95.2/84.8} & {93.8/90.8} & {99.2 / 96.0} & {99.7/89.4} & {98.6/93.2} & {97.2/92.4 }& {98.5/92.6 } \\
\midrule
Mean      & {76.8/85.6} & {84.2/89.5} & {81.9/84.9} & {88.1/87.2} & {85.1/88.9} & {96.4/95.7} & {93.4/96.0} & {92.6 / 95.6} & {96.5 / 96.8} & {98.0/97.3}& {96.3/96.6} \\
\bottomrule
\end{tabular}}
\end{table*}
\subsection{Ablation Study}
\begin{figure}
\setlength{\abovecaptionskip}{0pt}
\setlength{\belowcaptionskip}{0pt} 
\centering 
\includegraphics[width=\linewidth]{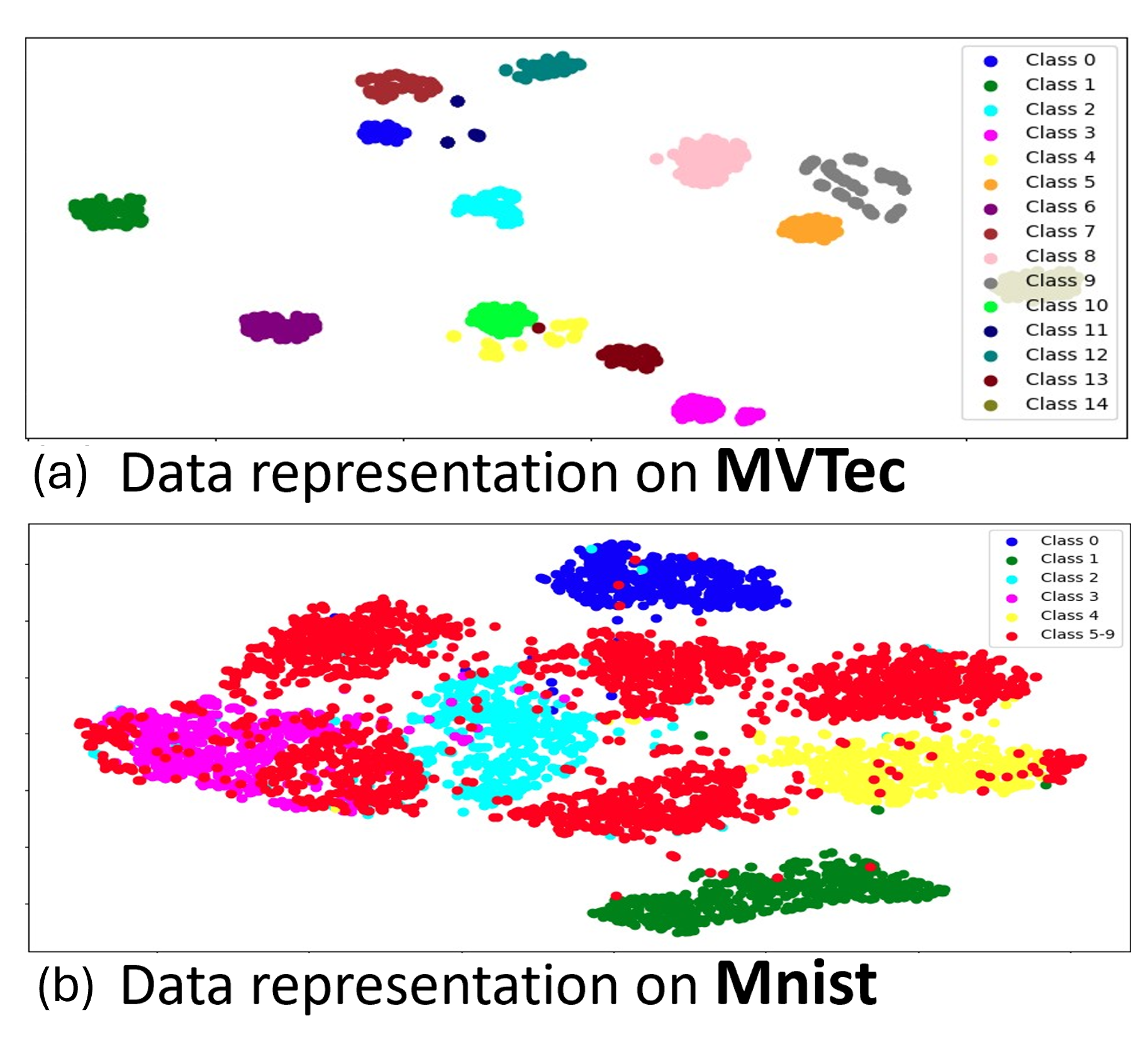}
   \caption{Visualization of Latent Features Across All Classes in MVTec and MNIST Datasets.(a) Each cluster represents a distinct class from the MVTec dataset, showing significant diversity in feature representation, indicative of unique features for each class.
(b) Clusters from the MNIST dataset exhibit a relatively compact distribution, indicating that each class has a less diverse feature representation. This visualization highlights the differences in feature variability between industrial and standard digit datasets.
 }
\label{feature_dis}
\end{figure}

To empirically validate the effectiveness of the proposed Feature Purification Module (FPM) and CFG decoder within MAD-ProFP, we conducted an ablation study. The results are summarized in Table~\ref{tab:ablation}.
We have observed that (i) without FPM, the performence drop drastically by 10\% to 19\%, which substantiate our assumption that restricting the latent space is critical to exclude anomalous embedding.(ii)CFG docder is only effective only with FPM module, which coincide with the purpose of CFG decoder by complementing the loosed information after features is filtered with FPM.
(iii) Within the FPM,manually set the top-k parameter is better than automatic filtered strategy. 
In the following passage, we will discuss the effectiveness of  each component separately:

\textbf{Analysis of the Feature Purification Module (FPM):}\\
As demonstrated in Table~\ref{tab:ablation}, the absence of the Feature Prototyping Module (FPM) leads to a significant decline in model performance: approximately 10/8\% (image/pixel level) on the industrial dataset and up to 19/13\% (CIFAR10/MNIST) on out-of-distribution (OOD) tasks. The FPM is specifically designed to condition the boundary of the learned distribution to exclusively align with the normality prototype embedding. This design is critical, particularly for OOD tasks, where embeddings of anomalous samples span across multiple classes of normal samples, as shown in Figure~\ref{feature_dis}. Without the FPM, these embeddings are more challenging to differentiate, leading to a more severe performance degradation. This observation underscores the unique and significant contribution of the FPM, distinguishing our approach from that proposed by You et al. \cite{you2022unified}.

Moreover, we initially implemented an automatic filtering mechanism in the FPM, as opposed to a manual top-k selection process. This mechanism masks features when the attention-based similarity between the prototype and the encoder feature falls below a predefined threshold (set at 0.5). However, empirical results indicate that the FPM with top-k selection outperforms the automatic mechanism by 1-2

\textbf{Analysis of the CFG Decoder:}\\
The CFG (Contextual Feature Generator) decoder exhibits differential efficacy across tasks, excelling in industrial anomaly detection, as evidenced by the MVTec dataset results, but showing a performance degradation in multi-class Out-of-Distribution (OOD) scenarios, particularly in the CIFAR10 and MNIST datasets. As shown in Table~\ref{tab:ablation}, while the CFG improves AUROC scores in MVTec from 94.3/96.1 to 96.3/96.6 (a gain of 2\% and 0.5\%, respectively), it results in a slight decrease in  MNIST performance.

In MVTec, the task is to identify subtle deviations within a homogeneous category set. Here, the reconstruction error between normal and anomalous samples is minimal, and the recovery of fine-grained features becomes crucial. The CFG decoder is designed to augment the filtered features by reintroducing some lost normal features, enhancing the model's sensitivity to minor anomalies.

Conversely, in multi-class OOD scenarios such as with the MNIST dataset, where feature diversity among classes is relatively low (as shown in Fig~\ref{feature_dis}. where the feature embedding among all classes in Mnist are more compact compared to MVTec), the skip connections in the CFG decoder may inadvertently facilitate the reconstruction of unseen anomalous classes. This could explain the observed slight drop in performance, as the CFG decoder may preserve features that contribute to false positives in anomaly detection, particularly in environments with less diverse and subtle class distinctions.
\begin{table}[ht]
\caption{The ablation study of FPM, and CFG decoder component.}
\label{tab:ablation}
\centering
\scriptsize
\begin{tabular}{lllllll}
\hline
\multicolumn{7}{c}{Component Ablation} \\
\hline
\multicolumn{1}{l|}{\multirow{2}{*}{w/o FPM}} & \multicolumn{2}{l|}{w/ FPM} & \multicolumn{1}{l|}{\multirow{2}{*}{CFG}} & \multicolumn{1}{l|}{Industry} & \multicolumn{2}{l}{OOD} \\ 
\cline{5-7} 
\multicolumn{1}{l|}{} & auto & \multicolumn{1}{l|}{Top-k} & \multicolumn{1}{l|}{} & MVTec & CIFAR10 & MNIST \\
\hline
\multicolumn{1}{l|}{v} & - & \multicolumn{1}{l|}{-} & \multicolumn{1}{l|}{-} & 85.6/88.3 & 73.2 & 75.4 \\
\multicolumn{1}{l|}{} & - & \multicolumn{1}{l|}{-} & \multicolumn{1}{l|}{v} & 85.5/88.8 & 75.4 & 73.4 \\
\multicolumn{1}{l|}{-} & v & \multicolumn{1}{l|}{} & \multicolumn{1}{l|}{v} & 95.6/96.4 & 92.5 & 88.5\\
\multicolumn{1}{l|}{-} & - & \multicolumn{1}{l|}{v} & \multicolumn{1}{l|}{} & 94.3/96.1 & 94.2 & 90.2\\
\multicolumn{1}{l|}{-} & - & \multicolumn{1}{l|}{v} & \multicolumn{1}{l|}{v} & 96.3/96.6 & 93.4 & 90.2 \\
\hline
\end{tabular}
\end{table}
\section{Discussion}
\subsection{Statistical Modeling of Reconstruction Boundaries in Relation to Dataset Variability}

\begin{table*}[]
\label{table:statistacl}
\centering
\caption{The probability density value of the anomalous embedding from the multi-class PDF is on average higher than the single-class PDF, indicating that the model trained on multi-classes tends to generalize covariate anomalies.}
\resizebox{\textwidth}{!}{%
\begin{tabular}{c|c|c|c|c|c|c|c|c|c|c|c|c|c|c|c}
\hline
\multicolumn{16}{c}{Probability density value on MVTec} \\
\hline
& Tile & Leather & Bottle & Grid & Transistor & Wood & Screw & Hazelnut & Cable & Zipper & Metal nuts & Pill & Capsule & Carpet & \textbf{Mean} \\
\hline
Single-class & 0 & 0 & 0 & 0 & 0 & 0 & 0 & 0 & 0 & 0 & 0 & 0 & 0 & 0 & \textbf{0} \\
\hline
Multi-class & 3.1 & 1.0 & 1.4 & 4.3 & 2.3 & 4.5 & 2.5 & 5.4 & 9.6 & 5.9 & 8.0 & 1.0 & 5.8 & 3.3 & \textbf{4.1} \\
\hline
\end{tabular}%
}
\label{table:statistacl1}
\end{table*}
\begin{table*}[]
\setlength{\abovecaptionskip}{0pt}
\setlength{\belowcaptionskip}{0pt}
\caption{The probability density value of the anomalous embedding from the multi-class PDF is on average higher than the single-class PDF, suggesting that a multi-class based model tends to have a more general boundary for anomalous sampling in the latent space, especially for semantic anomalies.}
\begin{center}
\begin{tabular}{c|cc|cc|cc}
\hline
\multicolumn{7}{c}{\textbf{Probability density value on MVTec}} \\ \hline
\multicolumn{1}{c|}{Class} & \multicolumn{2}{c|}{\{0,1,2,3,4\}} & \multicolumn{2}{c|}{\{1,3,5,7,9\}} & \multicolumn{2}{c}{\{0,2,4,6,8\}} \\ \cline{2-7} 
& Mnist & CIFAR10 & Mnist & CIFAR10 & Mnist & CIFAR10 \\ \hline
Single-class& 0 & 0 & 0 & 0 & 0 & 0 \\ \hline
Multi-class & 2.3 & 2.1 & 2.5 & 3.2 & 2.8 & 2.9 \\ \hline
\end{tabular}
\label{table:statistacl2}
\end{center}
\end{table*}
To further verify our assumption in Sec.~\ref{sec:Discussion of the behavior of reconstruction network in multi-class dataset},
We conduct experiment involving modeling the latent embeddings of both single-class and multi-class settings from the datasets using a Gaussian mixture model (GMM). This validate that anomlaous embedding are more likely to be interpreted as normal in Multi-class setting compared to single-class setting. 

As evidenced in Table.~\ref{table:statistacl1} and \ref{table:statistacl2}, The value of anomalous smaple from PDFs of multi-class embedding is much higher than PDFs of single class, which indicate the anomlaous sample prone to fall within the normality distribution in the learned latent space. 

the probability of anomlaous fall within the normality distribution in multiclass setting higher than  single-class with average 4.1 \% in MVTec dataset, 2.5\% in CIFAR10 and 2.5 \% in Mnist.

The observation substantiate our claim in Sec.~\ref{sec:Discussion of the behavior of reconstruction network in multi-class dataset}, which adding new category will boraden the variance of the latent variable distribution, which also generalize the reconstruction capability of the model to unseen anomaly, thereby potentially degrade the efficacy of anomaly detection performence.
\subsection{The Top-$k$ Parameter on Different Datasets}
\label{sec:topk}
In the FPM module, the pivotal hyper-parameter \(k\), which determines the number of embeddings retained from the encoder's representation, has a substantial influence on the performance outcomes, particularly in relation to the complexity of image compositions within different datasets.

For datasets like Mnist, which feature relatively simplistic image structures, a smaller top-$k$ suffices. The latest results, as shown in Table~\ref{tab:top-k}, indicate that Mnist achieves an AUROC score of 90.2\% with \(k=32\), supporting the use of fewer embeddings to effectively capture the dataset’s characteristics. Conversely, increasing \(k\) to 150 leads to a performance drop to an AUROC of 0.854, likely due to the unnecessary complexity added by excessive embeddings that may hinder the FPM module’s ability to effectively filter out anomalous structures.

In contrast, datasets such as MVTec and CIFAR10, characterized by their intricate image compositions, require a more considered approach to \(k\). MVTec achieves its best performance at \(k=150\) with an AUROC score of 96.6\%, while CIFAR10 reaches a peak AUROC score of 93.4\% at the same \(k\) value. These datasets benefit from retaining a larger number of embeddings, as a smaller top-$k$ might result in the loss of critical features, leading to significant reconstruction errors, affecting both anomalous and regular data identification.

For example, when \(k\) is set to 32, MVTec and CIFAR10 exhibit notably lower performance with AUROC scores of 91.3\% and 82.3\%, respectively. This indicates the potential for crucial embeddings being discarded, underscoring the necessity of adjusting \(k\) based on the specific requirements of the dataset.

This nuanced approach to selecting \(k\) is especially crucial in applications such as anomaly detection in CT scans, akin to the Mnist dataset, where choosing a smaller top-$k$, as demonstrated by the high performance at \(k=32\) for Mnist, could yield optimal results.

In summary, the choice of \(k\) must be tailored to the image complexity of the dataset, advocating against a one-size-fits-all approach. This insight presents a challenge for future work, aiming to develop a universal and effective strategy for selecting \(k\) across diverse datasets without manual adjustments.

\begin{table}[!htbp]
\setlength{\abovecaptionskip}{0pt}
\setlength{\belowcaptionskip}{0pt} 
\caption{The ablation study of different \(k\) variable on Mnist, MVtec, and CIFAR10 datasets.}
\centering
\begin{tabular}{clclclclcl}
\hline
\multicolumn{9}{|c|}{Ablation Study on $k$ variable}                                                                                                                                                                                                              \\ \hline
\multicolumn{2}{|c}{k=32} & \multicolumn{2}{c}{k=100} & \multicolumn{2}{c|}{k=150} & \multicolumn{1}{c|}{\begin{tabular}[c]{@{}c@{}}MVTec\\ score\end{tabular}} & \multicolumn{1}{c|}{\begin{tabular}[c]{@{}c@{}}Mnist\\ score\end{tabular}} & \multicolumn{1}{c|}{\begin{tabular}[c]{@{}c@{}}CIFAR10 \\ score\end{tabular}} \\ \hline
\multicolumn{2}{|c}{v}    & \multicolumn{2}{c}{}      & \multicolumn{2}{c|}{}      & \multicolumn{1}{c|}{\num{91.3}}                                                 & \multicolumn{1}{c|}{\num{90.2}}                                                  & \multicolumn{1}{c|}{\num{82.3}}    \\ \hline
\multicolumn{2}{|l}{}     & \multicolumn{2}{c}{v}     & \multicolumn{2}{l|}{}      & \multicolumn{1}{c|}{\num{96.1}}                                                & \multicolumn{1}{c|}{\num{88.5}}                                                  & \multicolumn{1}{c|}{\num{93.4}}    \\ \hline
\multicolumn{2}{|c}{}     & \multicolumn{2}{c}{}      & \multicolumn{2}{c|}{v}     & \multicolumn{1}{c|}{\num{96.6}}                                                 & \multicolumn{1}{c|}{\num{85.4}}                                                  & \multicolumn{1}{c|}{\num{93.0}}    \\ \hline
\end{tabular}
\label{tab:top-k}
\end{table}
\subsection{Visualization of Prototype Embeddings}

\textbf{Experimental Setting:}
To understand the multi-class prototype embeddings in our model, we visualize these embeddings using a pre-trained deconvolution network. This network reconstructs the high-dimensional latent prototype embeddings, $p^{i}_{\theta} \in \mathbb{R}^{s \times e}$, back into the input image space, $x \in \mathbb{R}^{H \times W \times C}$, allowing for an intuitive interpretation of the encoded semantic information.

\begin{figure*}
\setlength{\abovecaptionskip}{0pt}
\setlength{\belowcaptionskip}{0pt} 
\centering 
\includegraphics[width=\linewidth]{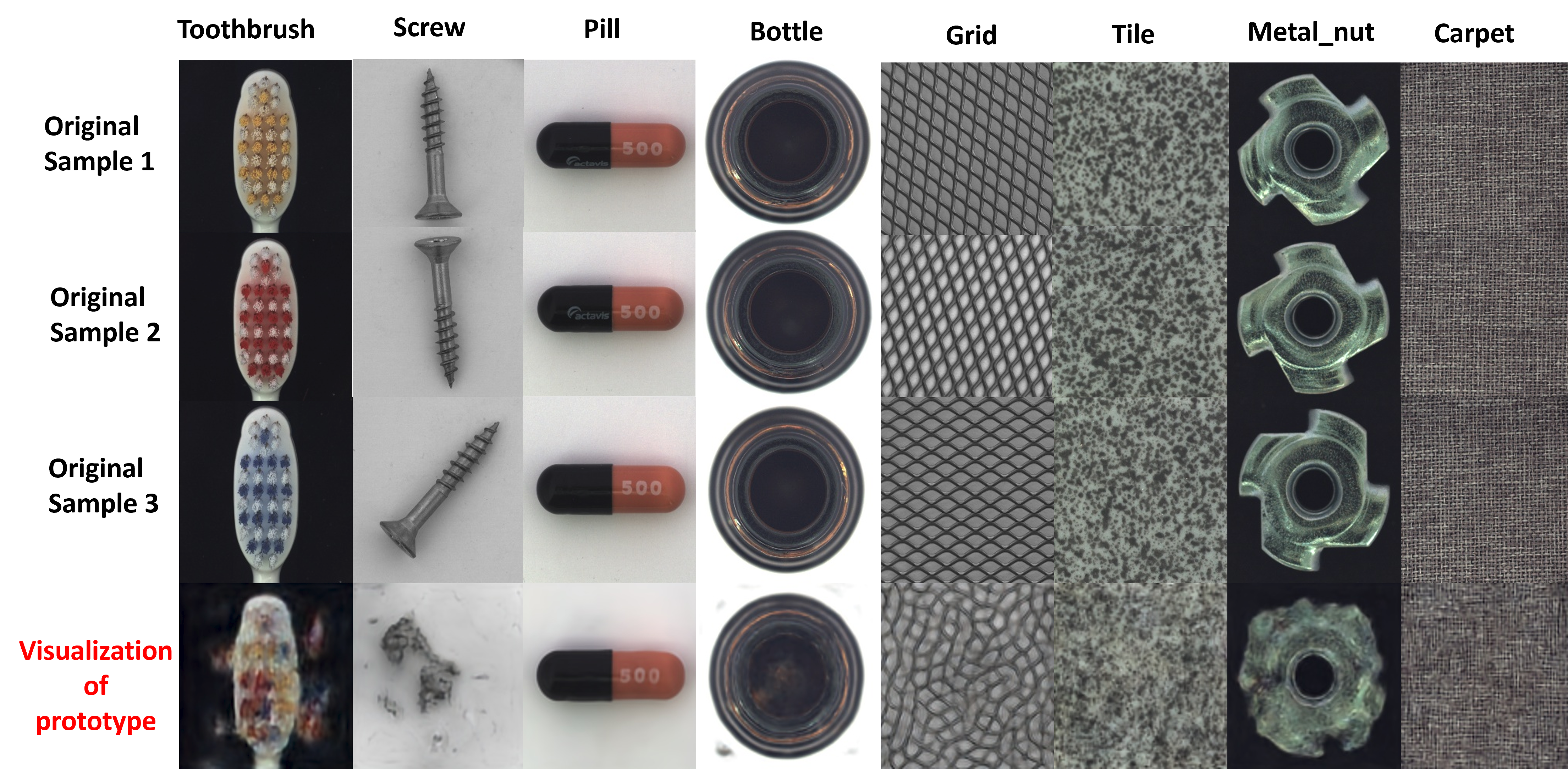}
   \caption{Visualization of Prototype Embeddings. The top rows show normal data samples from various MVTec classes, and the bottom row depicts corresponding prototype visualizations. These demonstrate the characteristic features captured by each prototype, guiding anomaly reconstruction within the MAD-ProFP framework.}
\label{vis_pro}
\end{figure*}

\textbf{The visualization result}
Visualizations of prototype embeddings (see Fig. \ref{vis_pro}) reveal distinct representations of class-specific features. Prototypes for classes with high intra-class variability (e.g., toothbrushes, screws, metal nuts) are more abstract, reflecting the diverse characteristics of the samples. In contrast, classes with more consistent features (e.g., bottles, pills, tiles) yield more precise and cohesive visualizations. These observations confirm that the prototypes effectively capture and represent the essential characteristics of each class, facilitating accurate anomaly detection through our reconstruction-based approach.

\section{Conclusion}
In this paper, we conduct a fundamental investigation into the underpinnings of poor performance in unsupervised anomaly detection across multi-class datasets. Our theoretical analysis identifies that an overly generalized boundary within the learned feature distribution in the latent space of deep networks is a primary factor contributing to these deficiencies. This insight led us to develop the FUTUREG framework, which constrains the expansive boundary of the latent feature space by anchoring to category-specific normality prototype embeddings and filtering out misaligned input embeddings.

Within the FUTUREG framework, the Feature Purification Module (FPM) plays a crucial role by filtering out anomalous embeddings based on the category-specific normality prototypes provided by the Normality Prototype Retrival Module (NPRM). Furthermore, the Cross-level Feature Guiding (CFG) decoder, an innovative adaptation of the transformer decoder, reconstructs the purified features. It utilizes a layer-wise encoder feature skip connection to reintroduce potentially filtered normal features, thereby enhancing the model’s reconstruction capabilities while continuing to exclude anomalous embeddings.

Our extensive experimental results validate the effectiveness of the proposed framework. FUTUREG demonstrates superior performance in multi-class Out-of-Distribution (OOD) detection compared to existing state-of-the-art methods and maintains competitive performance in industrial anomaly detection settings. This dual capability underscores the robustness and adaptability of our approach to various anomaly detection challenges.
\label{sec:conclusion}


{\small
\bibliographystyle{ieeenat_fullname}
\bibliography{11_references}
}

\end{document}